\documentclass[letterpaper, 10 pt, conference]{ieeeconf}

\IEEEoverridecommandlockouts

\usepackage{bbm}
\usepackage{etoolbox}
\usepackage{multicol}
\usepackage{cite}
\usepackage[bookmarks=true]{hyperref}
\usepackage{graphics}
\usepackage{times}
\usepackage{amsmath}
\usepackage{amssymb}
\usepackage{xspace}
\usepackage{mathrsfs}
\usepackage{amsfonts}
\usepackage{wrapfig}

\usepackage{kantlipsum}
\usepackage{subcaption}
\usepackage{marvosym}

\usepackage{graphicx}
\usepackage{float}
\usepackage{ctable}
\usepackage{cuted}
\usepackage{colortbl}
\usepackage{multirow}
\usepackage[misc,geometry]{ifsym}
\usepackage{algorithm}
\usepackage{algorithmic}
\usepackage{pifont}
\usepackage{array}
\usepackage{tabularx}
\usepackage{adjustbox}
\let\labelindent\relax
\usepackage{enumitem}
\usepackage{makecell}

\makeatletter
\let\NAT@parse\undefined
\makeatother

\definecolor{ADHeader}{RGB}{236,239,243}
\definecolor{ADDefault}{RGB}{228,239,251}

\usepackage{booktabs}

\newtheorem{theorem}{Theorem}

\newcommand{\kl}{\mathrm{kl}}

\newcommand{\Cset}{\mathcal C}
\title{Distinguish or Homogenize: Last-Chance Policy Identification and
Risk-Budgeted Recovery under Irreversible Resource Depletion}
\author{Yibo Guo$^{1}$, Xiaodan Wang$^{1}$\\
$^{1}$School of Computer and Artificial Intelligence, Zhengzhou University, Henan Zhengzhou 450001, China}

\begin{document}
\maketitle

\thispagestyle{empty}
\pagestyle{empty}

\begin{abstract}
Under irreversible resource depletion, an agent can spend resources to
distinguish among latent fault models, or to change the system state so that
the remaining models admit a common acceptable continuation---at which point
further diagnosis becomes unnecessary. This distinguish-or-homogenize principle
identifies a path that existing frameworks for identification, planning, and
diagnosis do not make explicit: prior formulations treat the mapping from fault
models to acceptable policies as a given, whereas LCPI makes it a function of
the agent's own actions. We formalize this principle through Last-Chance Policy
Identification (LCPI), where correctness is evaluated at the state the agent
reaches rather than at the initial state. The Last Identifiable Margin (LIM)
marks the feasibility boundary between distinguishing and homogenizing. For
deterministic diagnostic graphs we provide the Exact-LIM recursion; for noisy
finite-horizon recovery we propose Risk-Budgeted Compatibility Planning (RBCP),
which searches a compatibility-aware frontier under a hard worst-case failure
constraint. Across incident recovery on abstract microservice topologies and
latent-damage navigation in MiniGrid, RBCP improves risk-feasible recovery while
satisfying the failure budget. A sham control---cost-matched actions that
preserve model incompatibility---eliminates the gain entirely, confirming that
the benefit comes from changing which policies are acceptable for which models,
not from extra search or additional budget.
\end{abstract}

\section{Introduction}

A rescue robot is 200 meters underground. Its sensors return conflicting
readings: the signatures could indicate a methane leak ($\theta_1$) or smoke
from a structural fire ($\theta_2$). The battery reads 14\%. The robot can
run its gas chromatograph---costing an estimated 6\% battery---to determine
which fault is active. Or it can spend that same 6\% battery to force open
the ventilation doors. Ventilation clears the shaft regardless of whether
the gas is methane or smoke. The robot will not reach the surface if it
spends battery on both. It must choose.

The same structure recurs wherever diagnosis and intervention compete for
a single, irreversible resource pool. A cloud incident agent can spend its
CPU budget tracing root causes or rerouting traffic. A pilot handling an
engine warning can burn fuel on diagnostic checklists or on reaching an
alternate airport. In every case, the agent faces latent fault models, a
shrinking resource budget, and two ways to spend it: \textbf{distinguish}
which fault is active, or \textbf{homogenize} the situation so that the
remaining possibilities all admit the same recovery action.

Existing work offers three strategies, each visible in how the robot might
act. \textbf{Fixed-confidence identification} runs the chromatograph,
identifies $\theta_1$, and deploys the methane-specific ventilation
pattern---optimal if diagnosis finishes in time, catastrophic if it exhausts
the battery first. \textbf{Robust planning} deploys a single conservative
air-filtration protocol that works for both faults but never achieves the
efficiency of a targeted fix. \textbf{Decision-focused diagnosis} tests only
until the remaining fault models already share a recovery action---so if
methane and smoke happen to require the same scrubber, no further testing is
needed.

These three strategies differ in objective, but they share a structural
limit. None of them considers that an intervention can \textbf{change which
recovery actions are acceptable for which fault models}. Opening the
ventilation doors does not merely move the robot toward safety. It changes
the tunnel atmosphere. In ventilated air, both $\theta_1$ and $\theta_2$
admit the same continuation policy: ``exit through the main shaft.'' The
agent no longer needs to know which fault occurred. It spent resources not
to identify the fault, and not to find a one-size-fits-all plan, but to
\textbf{eliminate the decision-relevance of the remaining uncertainty}.

We formalize this choice as \textbf{Last-Chance Policy Identification
(LCPI)}. The agent operates under a finite, irreversible resource budget and
faces a finite set of latent fault models $\Theta$. It must, before
resources run out, deploy a \textbf{continuation policy} from a fixed
library $\Pi$---a complete recovery strategy from the reached state. Every
action, diagnostic or interventional, draws from the same budget.

\begin{figure*}[t]
    \centering
    \includegraphics[width=\linewidth]{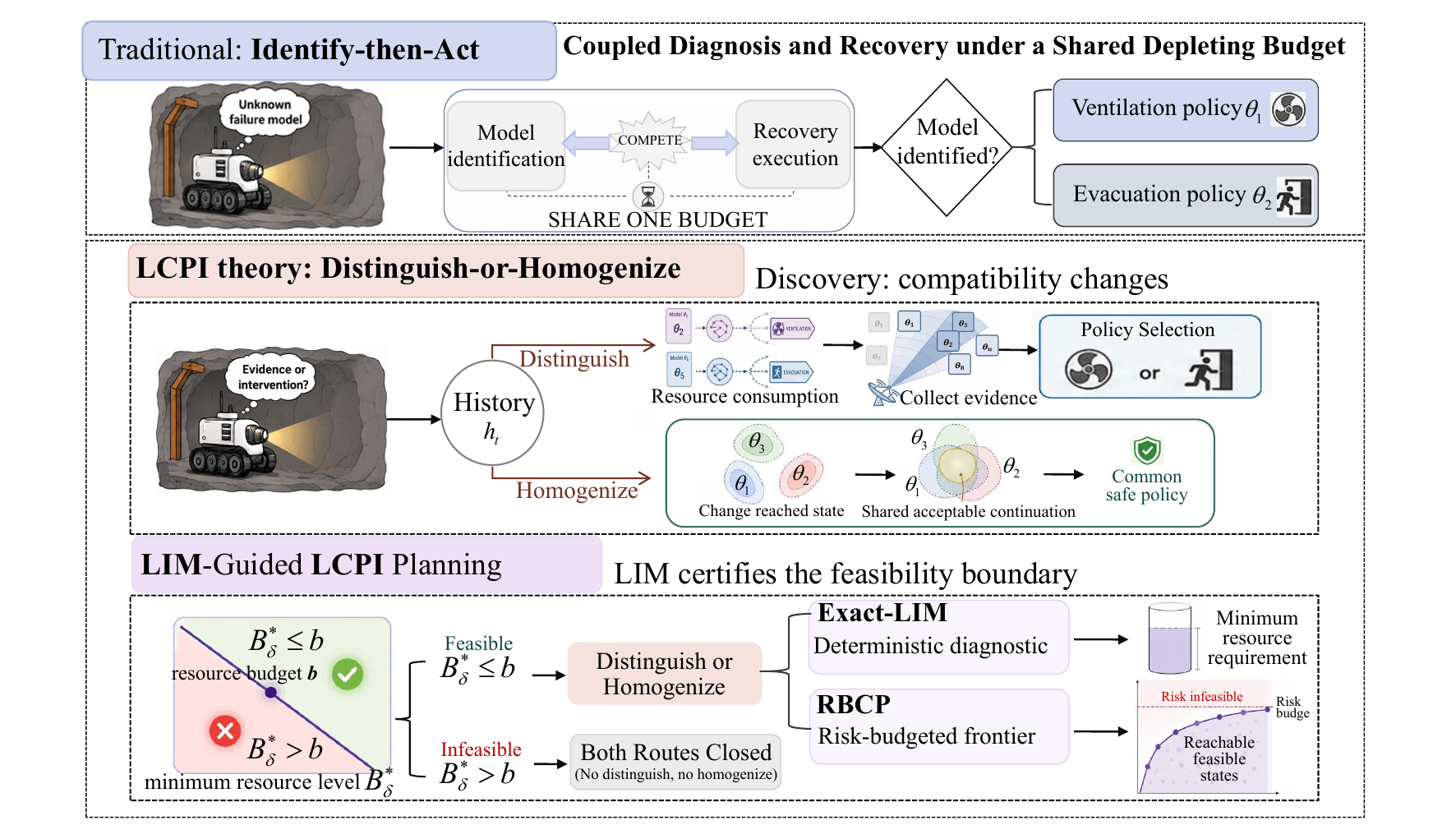}
    \caption{Overview of the LCPI framework. At each step the agent chooses
    between distinguishing active fault models and homogenizing the decision
    landscape. The Last Identifiable Margin (LIM) certifies the feasibility
    boundary between the two pathways.}
    \label{fig:lcpi_overview}
\end{figure*}

LCPI changes where correctness is evaluated. In the formulations underlying
fixed-confidence identification, robust planning, and decision-focused
diagnosis, a continuation policy is acceptable for model $\theta$ if it
performs well from the initial state. In LCPI, it is acceptable if it
performs well from \textbf{the state the agent reaches when it stops}. An
intervention thus does two things at once: it moves the agent toward
recovery, and it alters which policies count as acceptable for which
models. The mapping from models to acceptable policies---treated as a
given in prior problem formulations---becomes, in LCPI, a function of
the agent's own actions. Further diagnosis can then become unnecessary:
the agent does not know the fault, but it no longer needs to know.
Figure~\ref{fig:lcpi_overview} illustrates this architecture.

At every decision point, the agent faces a binary choice. It can spend
resources to gather evidence and narrow down which fault model is active
(\textbf{distinguish}), or spend resources to reach a state where all
remaining models admit a shared acceptable continuation
(\textbf{homogenize}). Both paths consume the same shrinking budget. Both
can lead to correct recovery. Choosing between them---and knowing when
each path remains viable---is the problem LCPI makes visible.

To decide whether distinguishing or homogenizing remains viable, we
introduce the \textbf{Last Identifiable Margin (LIM)}. Let $\mathcal
E^*(b)$ be the infimum worst-case probability of violating correctness
among all algorithms initialized with resource budget at least $b$, and
define $B_\delta^* = \inf\{\,b\ge 0 : \mathcal E^*(b) \le
\delta\,\}$. $B_\delta^*$ is the minimum initial resource level below
which no algorithm can guarantee correct recovery with confidence
$1-\delta$. Above LIM, either enough evidence can distinguish the
remaining models, or intervention can make them share a continuation.
Below LIM, neither path is guaranteed. LIM is a \textbf{feasibility
boundary for the distinguish-or-homogenize choice}, not a safety budget or
robust-planning objective.

Theorem~1 quantifies how the two paths trade off. Each model pair that
becomes compatible at the reached state reduces the probability mass on
which statistical identification must be performed. The total KL divergence
accumulated by the algorithm must exceed $\kl(1-\delta-g,\,\delta)$, where
$g$ is the probability mass on trajectories ending at a state where the
pair already shares a continuation. As $g$ increases, the required evidence
decreases monotonically; when $g\ge 1-2\delta$, the bound becomes zero.
\textbf{Homogenization reduces the probability mass requiring
identification, not the per-unit cost of gathering evidence.}

Two planning procedures operationalize this principle. On deterministic
diagnostic graphs, \textbf{Exact-LIM} gives an AND/OR recursion that
computes the minimum initial resource needed for zero-error recovery
(Theorem~2). On noisy finite-horizon domains, \textbf{Risk-Budgeted
Compatibility Planning (RBCP)} searches reachable states under resource
and risk budgets, prunes dominated plans, and deploys a certified fallback
only when all active models share an acceptable continuation while
satisfying the failure budget $\delta$.

Experiments use two controlled benchmarks: incident recovery on abstract
microservice topologies, and latent-damage navigation in MiniGrid. The
evaluation rule is \textbf{feasibility first}: any method that violates the
worst-case per-model risk budget is disqualified. RBCP clears this bar on
all test axes. Removing compatibility-changing actions drops WM-S-AUC by
0.12--0.34, and cost-matched sham actions that preserve incompatibility
produce the same drop. The gain comes from changing which policies are
acceptable for which models, not from additional actions or a larger search
budget.

The main contributions are as follows.
\begin{enumerate}
\item We identify the \textbf{distinguish-or-homogenize principle} as a
  discrete planning choice that is implicit in dual control, emergent in
  Bayes-Adaptive POMDPs, and absent from the formulations of
  decision-focused diagnosis---and make it explicit, with a computable
  feasibility boundary (LIM).
\item We formalize this principle through \textbf{LCPI}, where correctness
  is evaluated at the reached state, and prove a KL lower bound
  (Theorem~1) showing that homogenization reduces the probability mass
  that must be identified, not the per-unit cost of evidence.
\item We provide \textbf{Exact-LIM} for deterministic diagnostic graphs
  and \textbf{RBCP} for noisy finite-horizon recovery, and show via sham
  control that RBCP's gains come from compatibility-changing interventions,
  not from extra search or budget.
\end{enumerate}

\section{Related Work}

The introduction argued that fixed-confidence identification, robust
planning, and decision-focused diagnosis each treat the mapping from fault
models to acceptable policies as a given---a premise that leaves no room
for the distinguish-or-homogenize choice. The surrounding literature is
broad, so we focus on the specific point where each family stops short of
this choice.

\paragraph{Fixed-confidence and resource-limited identification.}
Fixed-confidence best-arm identification establishes lower bounds on the statistical evidence needed to identify the best arm with confidence $1-\delta$ \cite{garivier2016optimal}. Extensions to MDPs use active sampling or controlled trajectories to identify optimal or near-optimal policies \cite{almarjani2021navigating,wagenmaker2022pac}. Resource-aware variants add knapsack costs, per-arm budgets, or stopping rules under risk of ruin \cite{wang2022safe,li2023knapsack,vemulapati2026knapsacks,perotto2019ruin}. These methods assume a fixed decision criterion: the agent can optimize information acquisition, but cannot alter acceptability by changing the system state.

\paragraph{Decision-focused diagnosis and troubleshooting.}
Decision-focused diagnosis reduces wasted effort when several hypotheses imply the same action. Equivalence Class Determination groups hypotheses by shared decisions and tests only enough to separate groups \cite{golovin2010near}; decision-region determination generalizes this to overlapping regions \cite{javdani2014decision}; diagnosis-tree methods optimize test sequences by cost \cite{cicalese2014diagnosis}. Decision-theoretic troubleshooting interleaves observe, repair, and configure actions under one budget \cite{breese1996troubleshooting,stern2016batch}. However, configuration changes typically resolve uncertainty by removing faults, rather than making unresolved fault models share an acceptable continuation.

\paragraph{Safe planning and intervention under uncertainty.}
Dual control theory established that actions have both information and control value \cite{feldbaum1960dual}. Modern work adds shielding, minimax planning, and look-ahead. Viability theory characterizes states with at least one feasible continuation \cite{aubin1991viability}. Robust POMDPs, hidden-model POMDPs, and robust CMDPs optimize worst-case performance over model sets \cite{galesloot2025piprnn,galesloot2025rfpg,bovy2025mepomdp,kitamura2025rcmdp,ganguly2025rcmdp}. Constrained and recursively constrained POMDPs enforce risk budgets \cite{moss2024constrainedzero,ho2024rcpomdp}, and multi-cost reachability studies feasibility under multiple resource limits \cite{bork2025multicost}, and shielding for resource-constrained POMDPs prevents resource exhaustion through formal action blocking \cite{ajdarow2023shielding}. Across these methods, the model set and feasibility relations are planner inputs, not variables the planner can modify.

\paragraph{Positioning.}
None of these methods are wrong. Each solves the problem it was designed
for. Our claim is more specific: LCPI is, to our knowledge, the first
framework to (i)~explicitly isolate the distinguish-or-homogenize choice as
a discrete binary at each decision step, (ii)~provide a computable
feasibility boundary (LIM) for this choice under an irreversible resource
budget, and (iii)~quantify the resulting trade-off through a KL lower bound
(Theorem~1).

Prior frameworks capture related effects without making the choice itself
explicit. Dual control theory \cite{feldbaum1960dual} established that
actions carry both information and control value; modern descendants
balance probing against conservative regulation through a continuous
trade-off in a scalar cost function. The planner never explicitly decides
``now I distinguish, now I homogenize,'' and the framework provides no
analogue of the LIM---no computable threshold below which both paths close.
Bayes-Adaptive POMDPs \cite{ross2007bayes} augment the state with model
uncertainty, and actions that change the physical state can incidentally
make uncertainty decision-irrelevant. However, this homogenization effect
is emergent from the value function, not a planning objective with an
associated optimality certificate. Viability theory \cite{aubin1991viability}
characterizes the capture basin---the set of states from which a system can
reach a target while respecting constraints. The condition that all models
share an acceptable continuation is structurally analogous to reaching the
intersection of per-model viability kernels, but viability theory provides
no statistical identification cost and does not handle the discrete
diagnose-or-stabilize choice under a shared, irreversible budget.

LCPI makes compatibility a planning variable rather than a hidden
assumption. Exact-LIM and RBCP make the distinguish-or-homogenize choice
operational. Theorem~1 characterizes how the two paths trade off. Taken
together, these elements connect statistical identification,
viability-style feasibility analysis, and resource-constrained planning in
a way that no single prior framework does.

\section{The Distinguish-or-Homogenize Principle}

Return to the mine robot. It faces two fault models---gas leak or structural
fire---and must deploy a recovery plan before its battery dies. A recovery
plan is \textbf{acceptable} for a model if it performs well from the state
the robot reaches when it stops. At the start, the gas-specific scrubber
works for a leak but not a fire; the fire-suppression system works for a
fire but not a leak; the safe-exit corridor works for both. The robot can
spend battery on diagnostic checks (revealing which fault is active) or on
opening ventilation doors (changing the tunnel state so that both faults
admit the same recovery---the safe exit---without needing a diagnosis).

This is the distinguish-or-homogenize choice. At each step, the agent can
accumulate evidence to narrow down which fault is active
(\textbf{distinguish}), or change the state so that the remaining models
all share an acceptable recovery plan (\textbf{homogenize}). Both
paths consume the same finite, irreversible resource budget.

Not every state admits this choice. As resources run out, some fault
models lose all acceptable continuations---the point at which this happens
for the first model is the \textbf{loss-of-agency} moment. The agent must
stop before this moment and deploy a continuation policy. An algorithm is
\textbf{last-chance correct} if, with probability at least $1-\delta$, it
stops before loss-of-agency and deploys a continuation acceptable for the
true model from the reached state.

The \textbf{Last Identifiable Margin (LIM)} is the minimum initial resource
level below which no algorithm can meet this contract. Formally,
\[
B_\delta^* = \inf\{\,b\ge 0 : \mathcal E^*(b) \le \delta\,\},
\]
where $b$ denotes the initial resource budget, $\delta$ is the allowed
failure level, and $\mathcal E^*(b)$ is the smallest worst-case
probability of violating last-chance correctness that any feasible
algorithm can achieve when started with budget at least $b$. In other
words, $B_\delta^*$ is the minimum budget needed to make the task
feasible at confidence level $1-\delta$. When $b \ge B_\delta^*$, at
least one correct route remains open: the agent can either \emph{distinguish}
the active models by collecting enough diagnostic evidence, or
\emph{homogenize} them by reaching a state in which the remaining models
share the same acceptable continuation. When $b < B_\delta^*$, both
routes are closed, so no algorithm can guarantee correct recovery within
the remaining budget. The bottleneck is pairwise: correctness fails when
two models can be neither statistically distinguished by accumulating
enough diagnostic evidence nor made decision-compatible by reaching a
state where both accept the same continuation. Formal definitions, the
running example, and the LIM construction are in Supplement~\S S1.

\section{How Homogenization Reduces Identification Cost}

Why does reaching a shared continuation help? Suppose the robot stops at a
history where the gas leak and the fire already share an acceptable
recovery. On those trajectories, the robot no longer needs to tell the two
faults apart---the remaining uncertainty is decision-irrelevant. The more
probability mass lands on such consensus states, the less statistical
evidence the robot must accumulate elsewhere.

Theorem~\ref{thm:main} makes this quantitative. For any ordered pair of
fault models $(\theta,\theta')$, let $g_{\theta,\theta'}$ be the
probability, under $\theta$, that the algorithm stops at a history where
both models share an acceptable continuation. Let
$q_\theta(h,a)$ be the expected number of visits to history-action pair
$(h,a)$, and let $D_{\theta,\theta'}(h,a)$ be the KL divergence between the
observation distributions of $\theta$ and $\theta'$ at $(h,a)$.

\begin{theorem}[Residual Identification Cost under Reachable Consensus]
\label{thm:main}
Under finite-horizon and absolute-continuity assumptions, for every
$(\varepsilon,\delta)$-last-chance correct algorithm with $\delta<1/2$ and
every ordered model pair $(\theta,\theta')$,
\begin{equation}
\sum_{h,a} q_\theta(h,a) D_{\theta,\theta'}(h,a)
\ge
\phi_\delta(g_{\theta,\theta'}),
\end{equation}
where $\phi_\delta(g)=\kl(1-\delta-g,\,\delta)$ when $g<1-2\delta$, and
zero otherwise.
\end{theorem}

The bound tells a clean story. When no consensus state is reachable
($g_{\theta,\theta'}=0$), the required evidence equals the classical
fixed-confidence lower bound $\kl(1-\delta,\delta)$. As the consensus mass
grows, the required evidence decreases monotonically. When
$g_{\theta,\theta'}\ge 1-2\delta$, the bound reaches zero:
homogenization alone covers enough probability mass that no further
pairwise distinction is needed. Diagnosis and homogenization consume the
same budget but act on different quantities: diagnosis reduces uncertainty
about which model is active; homogenization reduces the decision impact of
the uncertainty that remains. Theorem~\ref{thm:main} quantifies the
trade-off. The full proof, via a stopped-history KL decomposition, is in
Supplement~\S S2.

\section{Compatibility-Aware Planning}

Theorem~\ref{thm:main} gives a design rule: favor actions that make
remaining models share acceptable continuations. We implement this rule
with two procedures.

\subsection{Exact-LIM: Deterministic Diagnostic Graphs}

On a deterministic diagnostic graph---a DAG where each diagnostic action
reveals which outcome occurred and consumes known resources---the minimum
resource budget required for zero-error recovery satisfies an AND/OR
recursion. At a node with version space $C$, if all models in $C$ already
share an acceptable continuation, the agent can deploy the cheapest one
immediately (the homogenize branch). Otherwise, it must select a diagnostic
action, observe the outcome, and recurse, paying the action cost plus the
worst-case cost over outcomes (the distinguish branch). The recursion
(proved in Supplement~\S S3) gives the exact minimum margin:

\[
B^*(C,v)=\min\Bigg\{
\begin{array}{l}
\displaystyle\min_{d\in\mathcal D(C,v)} r(v,d),\\[10pt]
\displaystyle\min_a\;\max_{o:C_o\neq\emptyset}\;
\bigl[c(v,a,o)+B^*(C_o,v_o)\bigr]
\end{array}
\Bigg\}.
\]

\subsection{RBCP: Noisy Finite-Horizon Recovery}

Real domains are noisier: observations are stochastic, transitions are
uncertain, and the version space rarely collapses cleanly.
\textbf{Risk-Budgeted Compatibility Planning (RBCP)} keeps the same
two-branch logic but replaces the exact AND/OR tree with a depth-$L$
frontier search over reachable states. Each frontier point records the
reached state, remaining resources, and per-model success and failure
estimates. The frontier is expanded exhaustively to depth $L$ and pruned
by coordinatewise dominance.

At each decision point, RBCP solves:
\[
\max_{\pi}\;\min_{\theta\in\Cset} S_\theta^\pi(h)
\quad\text{s.t.}\quad
\max_{\theta\in\Cset} F_\theta^\pi(h)\le\delta,
\]
where $\Cset$ is an anytime likelihood-ratio confidence set that contains
the true model with probability at least $1-\delta_c$. The constraint
enforces a hard per-model failure budget: every plausible model must
satisfy $F_\theta^\pi(h)\le\delta$. The objective maximizes worst-case
success among plans that clear this gate.

\begin{algorithm}[t]
\caption{Risk-Budgeted Compatibility Planning (RBCP)}
\label{alg:rbcp}
\begin{algorithmic}[1]
\REQUIRE $h_t$, reserve $\mathbf r_t$, likelihoods $\ell_t$, depth $L$, budget $\delta$, $\widehat K,\widehat V,\Pi$
\STATE $\Cset_t\gets\textsc{ConfidenceSet}(\ell_t)$
\STATE $(\mathcal F,b)\gets\textsc{InitFrontier}(h_t,\Cset_t,\Pi,\widehat V)$
\STATE $\mathcal F\gets\textsc{ExpandFrontier}(\mathcal F,L,\widehat K,\widehat V,\Pi)$
\STATE $\mathcal F\gets\textsc{PruneDominated}(\mathcal F)$
\STATE $\mathcal G\gets\{p\in\mathcal F:\max_{\theta\in\Cset_t}F_\theta(p)\le\delta\}$
\IF{$\mathcal G=\emptyset$ \AND $b\neq\bot$}
    \STATE \textbf{return} certified fallback $b$
\ELSIF{$\mathcal G=\emptyset$}
    \STATE \textbf{return} continuation minimizing $\max_{\theta\in\Cset_t}F_\theta$
\ENDIF
\STATE $p^*\gets\arg\max_{p\in\mathcal G}\min_{\theta\in\Cset_t}S_\theta(p)$
\STATE \textbf{return} policy encoded by $p^*$
\end{algorithmic}
\end{algorithm}

Algorithm~\ref{alg:rbcp} mirrors the deterministic recursion. If a
\textbf{certified fallback} exists---a continuation acceptable for every
model in $\Cset_t$ at the reached history---RBCP deploys it: that is
homogenization under noise. If the risk gate admits no plan, RBCP returns
the least risky available continuation.

Computing the exact LIM requires minimizing worst-case error over all
feasible algorithms---intractable on stochastic benchmarks. RBCP instead
instantiates the pairwise logic of Theorem~1 as a computable guard: for
each model pair in the confidence set, it estimates the resource cost to
either accumulate the required KL divergence (distinguish) or reach a
state where both models share a continuation (homogenize). If the maximum
over pairs exceeds the remaining usable budget, RBCP switches to the
certified fallback (Supplement~\S S6). The deployed plan's true failure
probability is at most $\delta_r+\delta_c$, where $\delta_r$ bounds the
estimated risk within the confidence set and $\delta_c$ bounds the
probability that the true model lies outside it.

\section{Experimental Evaluation}

The experiments test whether the mechanisms predicted by Sections~3--5 drive recovery performance. Three findings stand out. First, RBCP improves risk-feasible recovery on all test axes while satisfying the per-model failure budget (Q1). Second, the gain comes from compatibility-changing interventions, not from extra search or budget: a sham control that replaces these interventions with cost-matched actions preserving incompatibility removes the gain (Q3). Third, a scalar resource certificate is sufficient when resource direction is irrelevant, but the full resource vector matters when diagnostic chains consume heterogeneous reserves (Q4).

\textbf{Evaluation criterion.} A method is acceptable only if it satisfies the worst-mode risk constraint $\max_\theta F_\theta \le \delta$ with $\delta=0.10$. Among methods that pass this gate, we compare worst-mode recovery success. Methods that violate the risk budget are infeasible regardless of average performance. This criterion directly tests whether a planner respects the correctness contract required by LCPI.

\subsection{Tasks and Protocol}

We use two environments. Neither simulates a full deployment. Both isolate the distinguish-or-homogenize mechanism from competing explanations.

\paragraph{Microservice Incident Recovery (MIR).}
A cluster of interdependent services experiences a latent fault. Partial logs, traces, and health signals arrive on the service dependency graph. The agent can spend CPU, retry budget, and replica redundancy on diagnosis, such as tracing call chains or checking logs, or on intervention, such as rerouting traffic or failing over to a standby. Some interventions homogenize the model set: for example, failing over to a standby database works regardless of which service crashed. The agent can either identify the root cause or change the topology so that the remaining plausible causes accept the same recovery plan.

\paragraph{MiniGrid-LatentDamage (MLD).}
A robot moves through a grid with unobserved faults affecting sensors, actuators, doors, localization, and battery. Diagnostic actions, such as calibration and system checks, consume battery and can lead into irrecoverable regions. Recovery actions, such as taking a well-lit corridor or using backup navigation, allow multiple fault modes to share a viable path to the exit. A corridor that remains valid under both sensor and actuator faults homogenizes those models, because the robot does not need to know which fault occurred.

\paragraph{Design.}
Both environments rule out three alternative explanations. Latent models share observations, so the task cannot be solved by trivial identification. Diagnostic actions consume irreversible reserves, so more diagnosis is not always better. Selected interventions alter policy compatibility at the reached state, so we can test whether gains come from compatibility changes rather than from additional actions, deeper search, or scalarized risk tuning.

\paragraph{Protocol.}
Each environment is evaluated across deterministic-locked and stochastic reserve-shock settings at resource margins $\{2,\ldots,8\}$, using common random numbers on held-out instances. Protocol, seeds, search depths, test suites, and risk gates were frozen under source hash \texttt{b5bd13a5...d311} before execution, with strict development/locked-test separation. The instance is the statistical unit; we report paired bootstrap 95\% confidence intervals and paired sign-flip tests with Holm correction.

The resource-representation test (Q4) holds topology, observation models, transitions, policy values, and total diagnostic cost fixed. It compares balanced, time-rich, and redundancy-rich reserve profiles on disjoint held-out seeds. These profiles share the same scalar margin but differ in resource direction. Complete allocations and cell counts are given in Supplement~\S S7.

\subsection{Methods and Metrics}

\textbf{Baselines.} We compare RBCP against six alternatives that represent the main ways this problem could be framed: immediate fallback (pure rescue), diagnostic troubleshooting trees (DTT), equivalence-class determination (ECD), minimax POMDP planning, recursively constrained POMDP planning (RC-POMDP), and Lagrangian success--failure scalarization.

\textbf{Ablations.} Five ablations isolate the mechanisms predicted by the distinguish-or-homogenize principle:
\begin{itemize}
\setlength{\itemsep}{0pt}
\item \texttt{no\_hom}: removes all compatibility-changing actions.
\item \texttt{sham\_hom}: replaces compatibility-changing actions with cost-matched actions that preserve incompatibility. This is the decisive control, because any gain from compatibility change should disappear here.
\item \texttt{no\_lim}: removes LIM-based margin gating.
\item \texttt{no\_fallback}: removes the certified fallback.
\item \texttt{nonconditional}: removes branch-dependent continuations and forces one continuation for all active models.
\end{itemize}
The \texttt{scalar\_margin} ablation restricts the planner to the minimum reserve and plans from a balanced surrogate. Execution still uses the true resource vector, so this is a conservative information ablation rather than an unsafe relaxation.

\textbf{Metrics.} We report worst-mode task-success AUC (WM-S-AUC), pooled worst-mode failure (WM-F, evaluated at $\delta=0.10$), max-cell failure, $\varepsilon$-correctness, collapse rate, and online decision time.

\subsection{Q1: Does RBCP Improve Risk-Feasible Recovery?}

\begin{figure*}[t]
\centering
\includegraphics[width=0.94\textwidth]{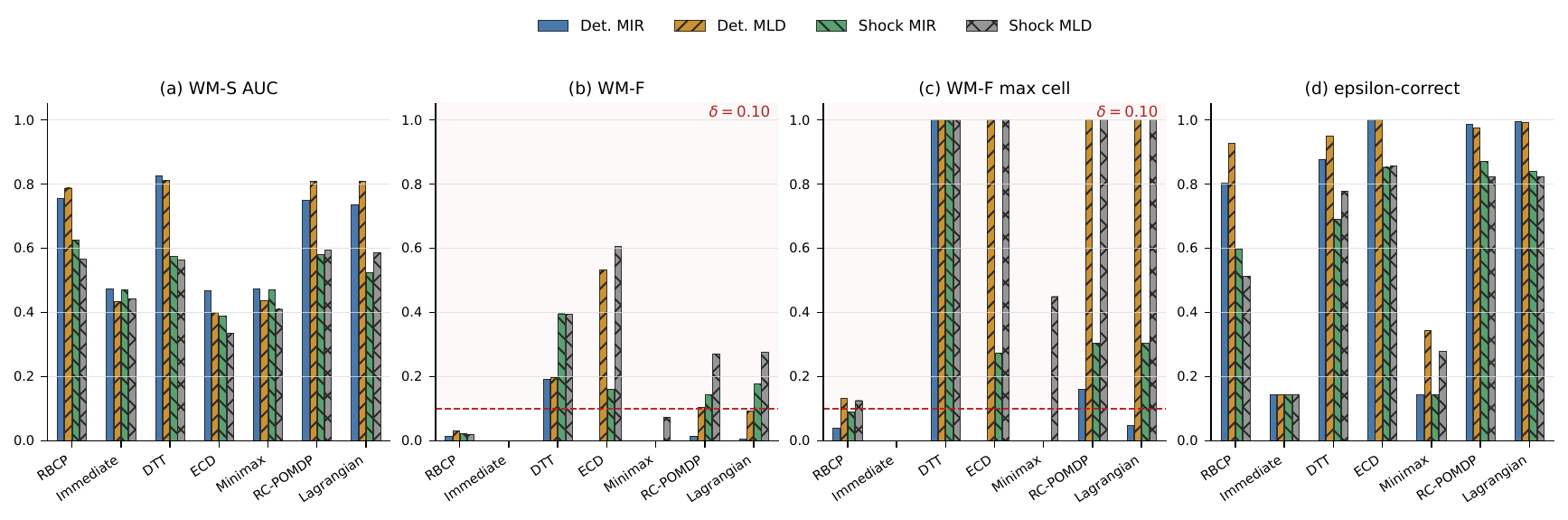}
\caption{Locked-test comparison across MIR/MLD deterministic and shock axes. Red lines mark $\delta=0.10$; methods above the line are infeasible regardless of success.}
\label{fig:locked-main-bars}
\end{figure*}

\begin{table}[t]
\centering
\scriptsize
\setlength{\tabcolsep}{2.7pt}
\begin{tabular}{llcccc}
\toprule
Axis & Env. & Comparison & Difference & 95\% CI & $p$ \\
\midrule
Det. & MIR & RBCP vs.\ RC-POMDP & +0.0067 & [0.0028,0.0101] & 0.0010 \\
Det. & MLD & RBCP vs.\ Minimax & +0.3512 & [0.3359,0.3673] & 0.0005 \\
Shock & MIR & RBCP vs.\ Minimax & +0.1553 & [0.1499,0.1607] & 0.0005 \\
Shock & MLD & RBCP vs.\ Minimax & +0.1558 & [0.1351,0.1776] & 0.0005 \\
\bottomrule
\end{tabular}
\caption{Locked-test paired comparisons in WM-S-AUC. Positive values favor RBCP. All comparisons pass the noninferiority gate.}
\label{tab:confirmatory}
\end{table}

Figure~\ref{fig:locked-main-bars} shows the main result. RBCP improves risk-feasible recovery on all four axes while satisfying the WM-F and max-cell failure budgets. Several alternatives achieve higher raw success on the MLD and shock axes, but they violate the risk budget and are therefore infeasible. This is the key LCPI distinction: RBCP does not maximize unconstrained success. It improves recovery under the per-model correctness contract $F_\theta \le \delta$. Table~\ref{tab:confirmatory} confirms this pattern. RBCP is never worse than the strongest feasible baseline on any axis and passes noninferiority everywhere, although it does not universally dominate because the best-feasible-method gate is stricter. RBCP requires 0.330~ms average online decision time on locked tests.

\subsection{Q2: Why Do Standard Planners Fall Short?}

The infeasibility patterns in Figure~\ref{fig:locked-main-bars} show that each baseline misses a different part of the LCPI contract. Fallback and minimax POMDPs stay below the risk budget, but their success is low because they never exploit compatibility-changing interventions. DTT and ECD achieve higher success on deterministic axes, but on shock axes their WM-F reaches 0.25--0.40 because they spend recovery margin on diagnosis without checking whether the reached state still supports a shared continuation. RC-POMDP remains feasible, but its recursive constraints are specified at the initial state and do not adapt to compatibility changes along the trajectory. Lagrangian scalarization attains the highest raw success on MLD-shock (AUC 0.91), but it violates WM-F at 0.32, because a single objective cannot enforce the per-model constraint required here. RBCP avoids these failure modes by verifying continuations after every reached-state compatibility change under a constraint that is both per-model and state-dependent. No baseline does both.

\subsection{Q3: Which RBCP Mechanisms Drive the Gain?}

\begin{figure*}[t]
\centering
\includegraphics[width=0.94\textwidth]{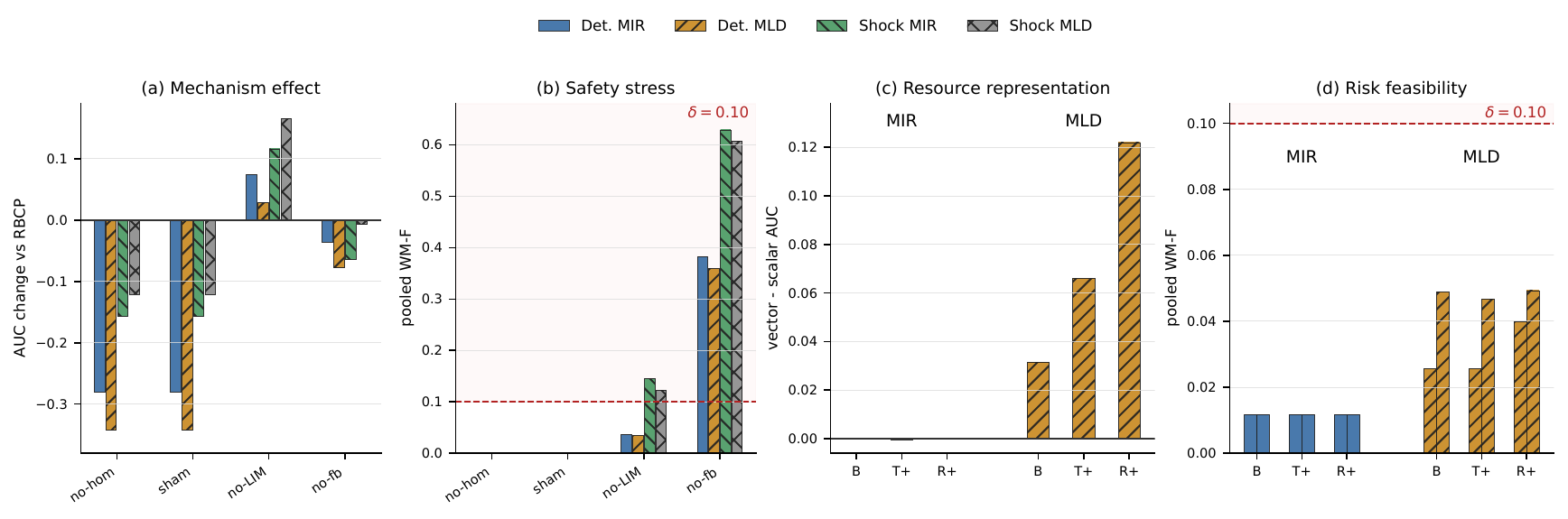}
\caption{Ablations for mechanism and resource representation. (a--b) WM-S-AUC change and pooled WM-F across the four locked-test axes. (c--d) Vector RBCP versus the scalar-margin ablation on balanced (B), time-rich (T+), and redundancy-rich (R+) profiles.}
\label{fig:locked-ablation-bars}
\end{figure*}

Figure~\ref{fig:locked-ablation-bars} shows that the gain is mechanism-specific, not incidental. Removing compatibility-changing actions (\texttt{no\_hom}) drops WM-S-AUC by 0.121--0.343, and the sham control (\texttt{sham\_hom}) matches that drop exactly. Because the two settings cost the same and expand the search tree equally, this match is the strongest evidence that the gain comes from changing policy compatibility, not from more actions or a larger search budget. Removing LIM-based gating (\texttt{no\_lim}) increases raw WM-S-AUC but pushes WM-F above 0.25 on shock axes, so aggressiveness comes at the cost of feasibility. Removing the certified fallback (\texttt{no\_fallback}) raises WM-F to 0.358--0.629, and removing branch-dependent continuations (\texttt{nonconditional}) lowers WM-S-AUC by 0.08--0.19. Together, these ablations show that compatibility change, margin gating, certified fallback, and branch-dependent continuation are all necessary components of the distinguish-or-homogenize strategy.
\subsection{Q4: When Does the Resource Vector Matter?}

This boundary test asks when the scalar LIM abstraction breaks down. LIM uses only a scalar resource level, but real resources are multi-dimensional, so a scalar summary discards direction. Figure~\ref{fig:locked-ablation-bars}c--d compares vector RBCP with the scalar-margin ablation on balanced (B), time-rich (T+), and redundancy-rich (R+) profiles. On MIR, the two methods perform almost identically, so resource direction does not matter. On MLD, keeping the full resource vector improves WM-S-AUC by 0.032, 0.066, and 0.122 across the three profiles, with Holm-adjusted $p=0.0012$ in all cases, while both representations remain below the failure budget. The asymmetry is expected: MIR consumes a relatively homogeneous pool of CPU and retry budget, so the minimum resource is enough, whereas MLD uses battery for sensing and time for movement, so the resources are not interchangeable. A planner that sees only the minimum cannot distinguish a battery-rich, time-poor state from the reverse. Vector planning costs more computation because it tracks a larger set of reachable resource states. Complete interval data and runtimes are reported in Supplement~\S S7.

\section{Discussion and Conclusion}

This paper proposed the distinguish-or-homogenize principle for settings where diagnosis and recovery compete for irreversible resources. We formalized LIM, proved a lower bound on the evidence needed when both paths remain viable, and developed Exact-LIM and RBCP as practical planners. The experiments show that the observed gain comes from the mechanism itself, not from extra search or budget.

The main implication is that the policy mapping in identification and robust planning is not fixed: intervention can reshape the decision problem itself. LIM characterizes when this choice is feasible, and the planners show that it is reachable in practice. The current limits are clear: the lower bound is necessary but not sufficient, Exact-LIM is exact only on deterministic acyclic graphs, and RBCP relies on estimated interfaces, finite search depth, and a hard risk gate. Extending the theory to richer graphs and learning interfaces online are the most direct next steps.
\bibliographystyle{IEEEtran}
\bibliography{references}

\end{document}